\pdfoutput=1

\documentclass[11pt]{article}

\usepackage[final]{EMNLP2023}

\usepackage{times}
\usepackage{latexsym}

\usepackage[T1]{fontenc}

\usepackage[utf8]{inputenc}

\usepackage{microtype}

\usepackage{inconsolata}
\usepackage{adjustbox} 
\usepackage{enumitem}
\usepackage{multirow}
\usepackage[utf8]{inputenc}
\usepackage{xcolor,colortbl}
\usepackage{microtype}
\usepackage{graphicx}
\usepackage{inconsolata}
\usepackage{booktabs, subcaption}
\title{Choose the Final Translation from NMT and LLM hypotheses Using MBR Decoding: HW-TSC’s Submission to the WMT24 General MT Shared Task}

\author{
  Zhanglin Wu, Daimeng Wei, Zongyao Li, Hengchao Shang, Jiaxin Guo,   \\
  \textbf{
    Shaojun Li, Zhiqiang Rao, Yuanchang Luo, Ning Xie, Hao Yang
  }\\
  Huawei Translation Service Center, Beijing, China\\
  \{wuzhanglin2,weidaimeng,lizongyao,shanghengchao,guojiaxin1,\\
 lishaojun18,raozhiqiang,nicolas.xie,yanghao30\}@huawei.com \\
  }

\begin{document}
\maketitle
\begin{abstract}
This paper presents the submission of Huawei Translate Services Center (HW-TSC) to the WMT24 general machine translation (MT) shared task, where we participate in the English to Chinese (en$\rightarrow$zh) language pair. Similar to previous years' work, we use training strategies such as regularized dropout, bidirectional training, data diversification, forward translation, back translation, alternated training, curriculum learning, and transductive ensemble learning to train the neural machine translation (NMT) model based on the deep Transformer-big architecture. The difference is that we also use continue pre-training, supervised fine-tuning, and contrastive preference optimization to train the large language model (LLM) based MT model. By using Minimum Bayesian risk (MBR) decoding to select the final translation from multiple hypotheses for NMT and LLM-based MT models, our submission receives competitive results in the final evaluation.
\end{abstract}

\section{Introduction}

Machine translation (MT) \cite{brown1990statistical} predominantly utilizes transformer encoder-decoder architectures \cite{vaswani2017attention}, which is evident in prominent models such as NLLB-200 \cite{costa2022no}, M2M100 \cite{fan2021beyond}, and MT5 \cite{xue2021mt5}. Significant research effort has been devoted to task-specific neural machine translation (NMT) models \cite{wei2022hw,wu2023path} trained in a fully supervised manner with large volumes of parallel data. Their performance has been enhanced through techniques such as regularized dropout \cite{wu2021r}, bidirectional training \cite{ding2021improving}, data diversification \cite{nguyen2020data}, forward translation \cite{abdulmumin2021enhanced}, back translation \cite{sennrich2016improving}, alternated training \cite{jiao2021alternated}, curriculum learning \cite{zhang2019curriculum}, and transductive ensemble learning \cite{wang2020transductive}. 

The emergence of decoder-only large language models (LLMs) such as the GPT series \cite{wu2023brief,achiam2023gpt}, Mistral \cite{jiang2023mistral}, and LLaMA \cite{touvron2023llama,touvron2023llama2} shows remarkable efficacy in various NLP tasks, providing a fresh perspective on the MT task. Recent studies \cite{hendy2023good,jiao2023parrot} indicate that larger LLMs such as GPT-3.5 (175B) and GPT-4 exhibit strong translation abilities. However, the performance of smaller-sized LLMs (7B or 13B) still falls short when compared to conventional NMT models \cite{zhu2024multilingual}. Therefore, there are studies \cite{yang2023bigtranslate,zeng2024teaching} intend to enhance the translation performance for these smaller LLMs, but their improvements are relatively modest, primarily due to the predominant pre-training of LLMs on English-centric datasets, resulting in limited linguistic diversity. Addressing this limitation, Xu et al. \cite{xu2023paradigm} initially continue pre-training (CPT) LLaMA-2 \cite{touvron2023llama2} with extensive non-English monolingual data to enhance their multilingual abilities, and then perform supervised fine-tuning (SFT) with high-quality parallel data to instruct the model to generate translations. Nonetheless, the performance still lags behind leading translation models such as GPT-4 and WMT competition winners. Subsequently, Xu et al. \cite{xu2024contrastive} bridged this gap by further fine-tuning the LLM-based MT model using contrast preference optimization (CPO).

Ensembling \cite{zhou2002ensembling} has a long history in machine learning, being well known for leveraging multiple complementary systems to improve performance on a given task and provide good/robust generalization. Minimum Bayesian risk (MBR) \cite{finkelstein2023mbr,farinhas2023empirical} decoding has successfully improved translation quality using task-specific NMT models, and subsequently it has also been shown to be suitable for LLM-based MT models.

\begin{figure*}[ht] 
\centering
\includegraphics[width=120mm]{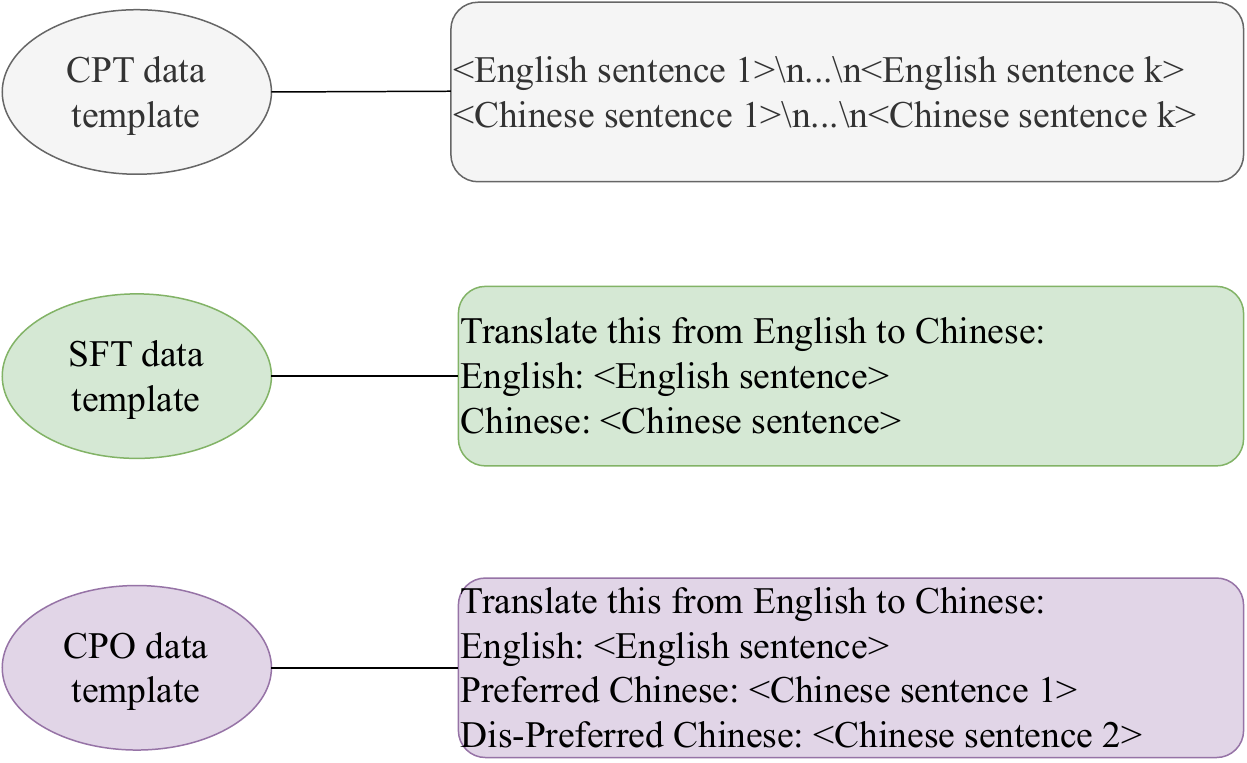}
\caption{\centering CPT, SFT and CPO data templates used for LLM-based MT training.}
\label{prompt}
\end{figure*}

For the WMT24 general MT shared task, we participate in the en$\rightarrow$zh language pair. Similar to previous years' work \cite{wei2021hw,wei2022hw,wu2023path}, we use training strategies such as regularized dropout \cite{wu2021r}, bidirectional training \cite{ding2021improving}, data diversification \cite{nguyen2020data}, forward translation \cite{abdulmumin2021enhanced}, back translation \cite{sennrich2016improving}, alternated training \cite{jiao2021alternated}, curriculum learning \cite{zhang2019curriculum}, and transductive ensemble learning \cite{wang2020transductive} to train NMT models based on the deep transformer-big architecture. In addition, we use CPT, SFT and CPO methods to train LLM-based MT models. Finally, we use MBR decoding to select the final translation from multiple hypotheses of NMT and LLM-based MT models.

\section{Data}

\subsection{Data Source} 

We obtain bilingual and monolingual data from ParaCrawl v9, News Commentary v18.1, Wiki Titles v3, UN Parallel Corpus V1.0, CCMT Corpus, WikiMatrix, News Crawl and Common Crawl data sources. The amount of data we used for training NMT and LLM-based MT models is shown in Table \ref{data}. It should be noted that in order to obtain better translation performance in the general domain, we mix the monolingual data from Common Crawl and News Crawl.

\begin{table}[ht]
\begin{center}
\begin{adjustbox}{width=\columnwidth,center}
\begin{tabular}{@{}ccc@{}}
\hline
language pairs &  bitext data & monolingual data \\
\hline
en$\rightarrow$zh & 25M & en: 50M, zh: 50M\\
\hline
\end{tabular}
\end{adjustbox}
\caption{Bilingual and monolingual used for training NMT and LLM-based MT models.}\label{data}
\end{center}
\end{table}

\subsection{NMT Data Pre-processing}

Our data pre-processing methods for NMT include: 
\begin{itemize}
    \item Remove duplicate sentences or sentence pairs.
    \item Convert full-width symbols to half-width.
    \item Use fasttext\footnote{\url{https://github.com/facebookresearch/fastText}} \cite{joulin2016fasttext} to filter other language sentences.
    \item Use jieba\footnote{\url{https://github.com/fxsjy/jieba}} to pre-segment Chinese sentences.
    \item Use mosesdecoder\footnote{\url{https://github.com/moses-smt/mosesdecoder}} \cite{koehn-etal-2007-moses} to normalize English punctuation.
    \item Filter out sentences with more than 150 words.
    \item Use fast-align \cite{dyer2013simple} to filter sentence pairs with poor alignment.
    \item Sentencepiece\footnote{\url{https://github.com/google/sentencepiece}} (SPM) \cite{kudo2018sentencepiece} is used to perform subword segmentation, and the vocabulary size is set to 32K.
\end{itemize}

Since there may be some semantically dissimilar sentence pairs in bilingual data, we use LaBSE\footnote{\url{https://huggingface.co/sentence-transformers/LaBSE}} \cite{feng2022language} to calculate the semantic similarity of each bilingual sentence pair, and exclude bilingual sentence pairs with a similarity score lower than 0.7 from our training corpus.

\subsection{LLM-based MT Data Pre-processing}

The training of the LLM-based MT model requires three stages: CPT, SFT and CPO. As shown in Figure \ref{prompt}, the training data templates of the LLM-based MT model in these three stages are different.


In the CPT stage, considering that most LLMs are trained on English-dominated data, we using Chinese and English monolinguals for CPT to improve LLM's proficiency in Chinese. To preserve the long-context modeling capability of LLM, we concatenate multiple sentences into a long text with no more than 4096 words, and preferentially concatenate sentences from the same document.

In the SFT stage, drawing inspiration from the recognized significance of data quality in other applications \cite{zhou2024lima,maillard2023small}, we fine-tune the model with high-quality parallel data. In order to obtain high-quality parallel data, we use cometkiwi model \footnote{\url{https://huggingface.co/Unbabel/wmt22-cometkiwi-da}} \cite{rei2022cometkiwi} to calculate the score of bilingual data on the en$\rightarrow$zh language pair, and then retain bilingual data with a cometkiwi score greater than 0.8.

In the CPO stage, to learn an objective that fosters superior translations and rejects inferior ones, access to labeled preference data is essential, yet such data is scarce in machine translation. The following describes our process of constructing the triplet preference data required for CPO training. First, we randomly sample 50,000 data from high-quality bilingual data. Then, we use the NMT model to obtain N-best (N=10) hypotheses based on beam search decoding, and then use the comet-da model\footnote{\url{https://huggingface.co/Unbabel/wmt20-comet-da}} \cite{rei2020comet} to calculate the score of each hypothesis, select the hypothesis with the highest score as the preferred translation, and select the hypothesis with the lowest score as the dis-preferred translation.

\section{NMT System}

\subsection{System Overview}

Transformer is the state-of-the-art model structure in recent NMT evaluations. There are two parts of research to improve this kind: the first part uses wide networks (eg: Transformer-Big \cite{vaswani2017attention}), and the other part uses deeper language representations (eg: Deep Transformer \cite{wang2019learning}). For the WMT24 general MT shared task, we combine these two improvements, adopting the Deep Transformer-Big \cite{wei2022hw,wu2023path} model structure to train the NMT system. Deep Transformer-Big uses pre-layer normalization, features 25-layer encoder, 6-layer decoder, 16-heads self-attention, 1024-dimensional word embedding and 4096-dimensional FFN embedding.

Fig. \ref{Bilingual_Training} shows the overall training flow of NMT system. We use training strategies such as regularized dropout (R-Drop) \cite{wu2021r}, bidirectional training (BiT) \cite{ding2021improving}, data diversification (DD) \cite{nguyen2020data}, forward translation {FT) \cite{abdulmumin2021enhanced}, back translation (BT) \cite{sennrich2016improving}, alternated training (AT) \cite{jiao2021alternated}, curriculum learning (CL) \cite{zhang2019curriculum}, and transductive ensemble learning (TEL) \cite{wang2020transductive} for training.

\begin{figure}[t] 
\centering
\includegraphics[width=60mm]{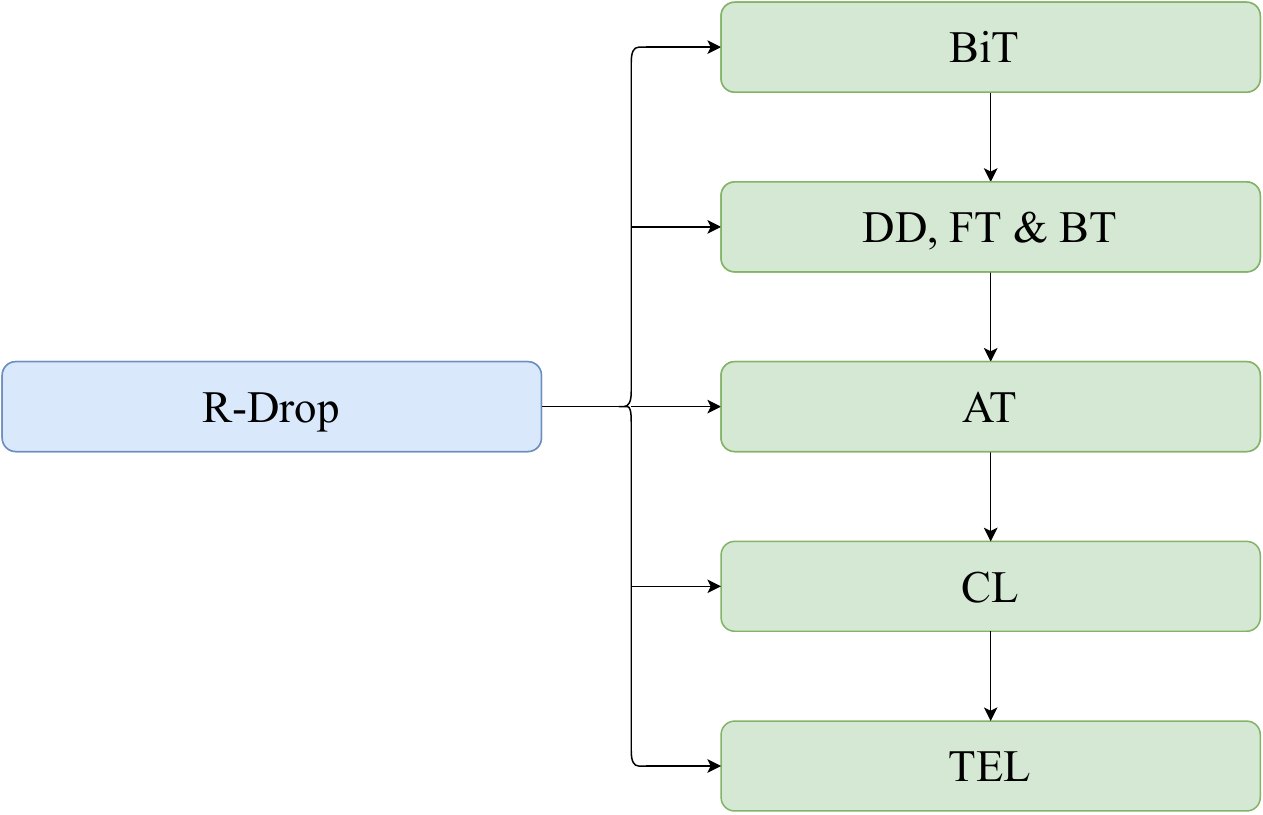}
\caption{\centering The overall training flow of NMT system.}
\label{Bilingual_Training}
\end{figure}

\subsection{Regularized Dropout}

Regularized Dropout (R-Drop)\footnote{\url{https://github.com/dropreg/R-Drop}} \cite{wu2021r} is a simple yet more effective alternative to regularize the training inconsistency induced by dropout \cite{srivastava2014dropout}. Concretely, in each mini-batch training, each data sample goes through the forward pass twice, and each pass is processed by a different sub model by randomly dropping out some hidden units. R-Drop forces the two distributions for the same data sample outputted by the two sub models to be consistent with each other, through minimizing the bidirectional Kullback-Leibler (KL) divergence \cite{van2014renyi} between the two distributions. That is, R-Drop regularizes the outputs of two sub models randomly sampled from dropout for each data sample in training. In this way, the inconsistency between the training and inference stage can be alleviated.

\subsection{Bidirectional Training}

Many studies have shown that pre-training can transfer the knowledge and data distribution, hence improving the model generalization. Bidirectional training (BiT) \citep{ding2021improving} is a simple and effective pre-training method for NMT. Bidirectional training is divided into two stages: (1) bidirectionally updates model parameters, and (2) tune the model. To achieve bidirectional updating, we only need to reconstruct the training samples from "src$\rightarrow$tgt" to "src$\rightarrow$tgt \& tgt$\rightarrow$src" without any complicated model modifications. Notably, BiT does not require additional parameters or training steps and only uses parallel data.

\subsection{Data Diversification}

Data Diversification (DD) \citep{nguyen2020data} is a data augmentation method to boost NMT performance. It diversifies the training data by using the predictions of multiple forward and backward models and then merging them with the original dataset which the final NMT model is trained on. DD is applicable to all NMT models. It does not require extra monolingual data, nor does it add more parameters. To conserve training resources, we only use one forward model and one backward model to diversify the training data.

\subsection{Forward Translation}

Forward translation (FT) \cite{abdulmumin2021enhanced}, also known as self-training, is one of the most commonly used data augmentation methods. FT has proven effective for improving NMT performance by augmenting model training with synthetic parallel data. Generally, FT is performed in three steps: (1) randomly sample a subset from the large-scale source monolingual data; (2) use a “teacher” NMT model to translate the subset data into the target language to construct the synthetic parallel data; (3) combine the synthetic and authentic parallel data to train a “student” NMT model.

\subsection{Back Translation}

An effective method to improve NMT with target monolingual data is to augment the parallel training data with back translation (BT) \cite{sennrich2016improving,wei-etal-2023-text}. There are many works expand the understanding of BT and investigates a number of methods to generate synthetic source sentences. \citeauthor{edunov2018understanding} (\citeyear{edunov2018understanding}) find that back translations obtained via sampling or noised beam outputs are more effective than back translations generated by beam or greedy search in most scenarios. \citeauthor{caswell2019tagged} (\citeyear{caswell2019tagged}) show that the main role of such noised beam outputs is not to diversify the source side, but simply to tell the model that the given source is synthetic. Therefore, they propose a simpler alternative strategy: Tagged BT. This method uses an extra token to mark back translated source sentences, which generally outperforms noised BT \cite{edunov2018understanding}. For better joint use with FT, we use sampling back translation (ST) \cite{edunov2018understanding}.

\subsection{Alternated Training}

While synthetic bilingual data have demonstrated their effectiveness in NMT, adding more synthetic data often deteriorates translation performance since the synthetic data inevitably contains noise and erroneous translations. Alternated training (AT) \cite{jiao2021alternated} introduce authentic data as guidance to prevent the training of NMT models from being disturbed by noisy synthetic data. AT describes the synthetic and authentic data as two types of different approximations for the distribution of infinite authentic data, and its basic idea is to alternate synthetic and authentic data iteratively during training until the model converges.

\subsection{Curriculum Learning}

A practical curriculum learning (CL) \cite{zhang2019curriculum} method should address two main questions: how to rank the training examples, and how to modify the sampling procedure based on this ranking. For ranking, we choose to estimate the difficulty of training samples according to their domain feature \cite{wang2020learning}. The calculation formula of domain feature is as follows, where $\theta_{in}$ represents an in-domain NMT model, and $\theta_{out}$ represents a out-of-domain NMT model. One thing to note is that we treat domains including news, user-generated (social), conversational, and e-commerce domains as in-domain, and others as out-of-domain. Specifically, we use the WMT22 test set to fine-tune a baseline model, and then use the baseline model and the fine-tuned model as the out-of-domain model and the in-domain model respectively.

\begin{equation}
q(x,y)=\frac{\log{P(y\vert x;\theta_{in})}-\log{P(y\vert x;\theta_{out})}}{\vert y\vert} \label{eq1}
\end{equation}

For sampling, we adopt a probabilistic CL strategy that leverages the concept of CL in a nondeterministic fashion without discarding the original standard training practice, such as bucketing and mini-batching.

\subsection{Transductive Ensemble Learning}

Ensemble learning \cite{garmash2016ensemble}, which aggregates multiple diverse models for inference, is a common practice to improve the performance of machine learning models. However, it has been observed that the conventional ensemble methods only bring marginal improvement for NMT when individual models are strong or there are a large number of individual models. Transductive Ensemble Learning (TEL) \cite{zhang2019curriculum} studies how to effectively aggregate multiple NMT models under the transductive setting where the source sentences of the test set are known. TEL uses all individual models to translate the source test set into the target language space and then finetune a strong model on the translated synthetic data, which significantly boosts strong individual models and benefits a lot from more individual models.

\section{LLM-based MT System}

\subsection{System Overview}


There is recently a surge in research interests in Transformer-based LLMs, such as ChatGPT \cite{wu2023brief}, GPT-4 \cite{achiam2023gpt}, and LLaMA \cite{touvron2023llama,touvron2023llama2}. Benefiting from the giant model size and oceans of training data, LLMs can understand better the language structures and semantic meanings behind raw text, thereby showing excellent performance in a wide range of natural language processing (NLP) tasks. Although the training methodology of LLMs is simple, high computational requirements have
limited the development of LLMs to a few players. In order to avoid training LLM from scratch, we chose to conduct research work on the open source Llama2-13b\footnote{\url{https://huggingface.co/meta-llama/Llama-2-13b-hf}} \cite{touvron2023llama2} model. Llama2-13b is an autoregressive language model using an optimized transformer architecture that is pre-trained on 2 trillion tokens of data from publicly available sources. As shown in Figure 3, we train Llama2-13b into a powerful LLM-based MT model through three-stage training of CPT, SFT and CPO. 

\begin{figure}[t] 
\centering
\includegraphics[width=60mm]{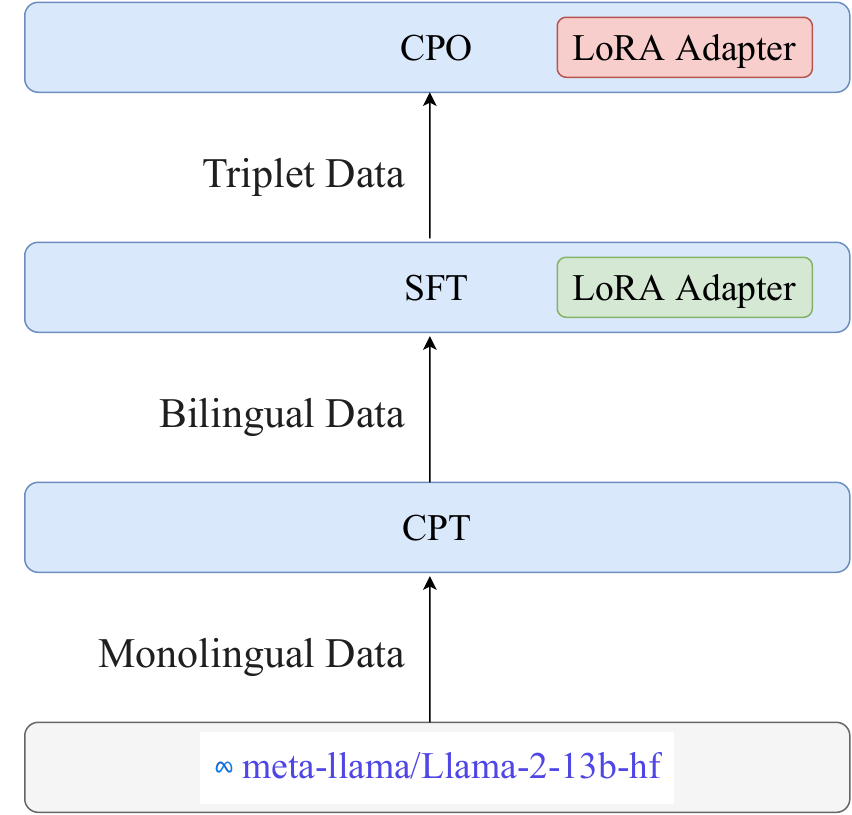}
\caption{\centering The training flow of LLM-based MT system.}
\label{LLM}
\end{figure}

\subsection{Continue Pre-training}

LLMs like LLaMA are pre-trained on English-dominated corpora. This potentially explains their inadequate translation performance which necessitates cross-lingual capabilities. To ameliorate this, our first stage is to perform continue pre-training (CPT) on LLM with Chinese and English monolingual data to improve proficiency in Chinese and prevent forgetting of English knowledge. Previous studies also offer some clues that monolingual data help in translation. For instance, guo et al. \cite{guo2024novel} proposed a three-stage training method, which proved that using CPT can improve the performance of MT task in the SFT stage. Note that we use full fine-tuning at this stage.

\begin{figure*}[ht] 
\centering
\includegraphics[width=120mm]{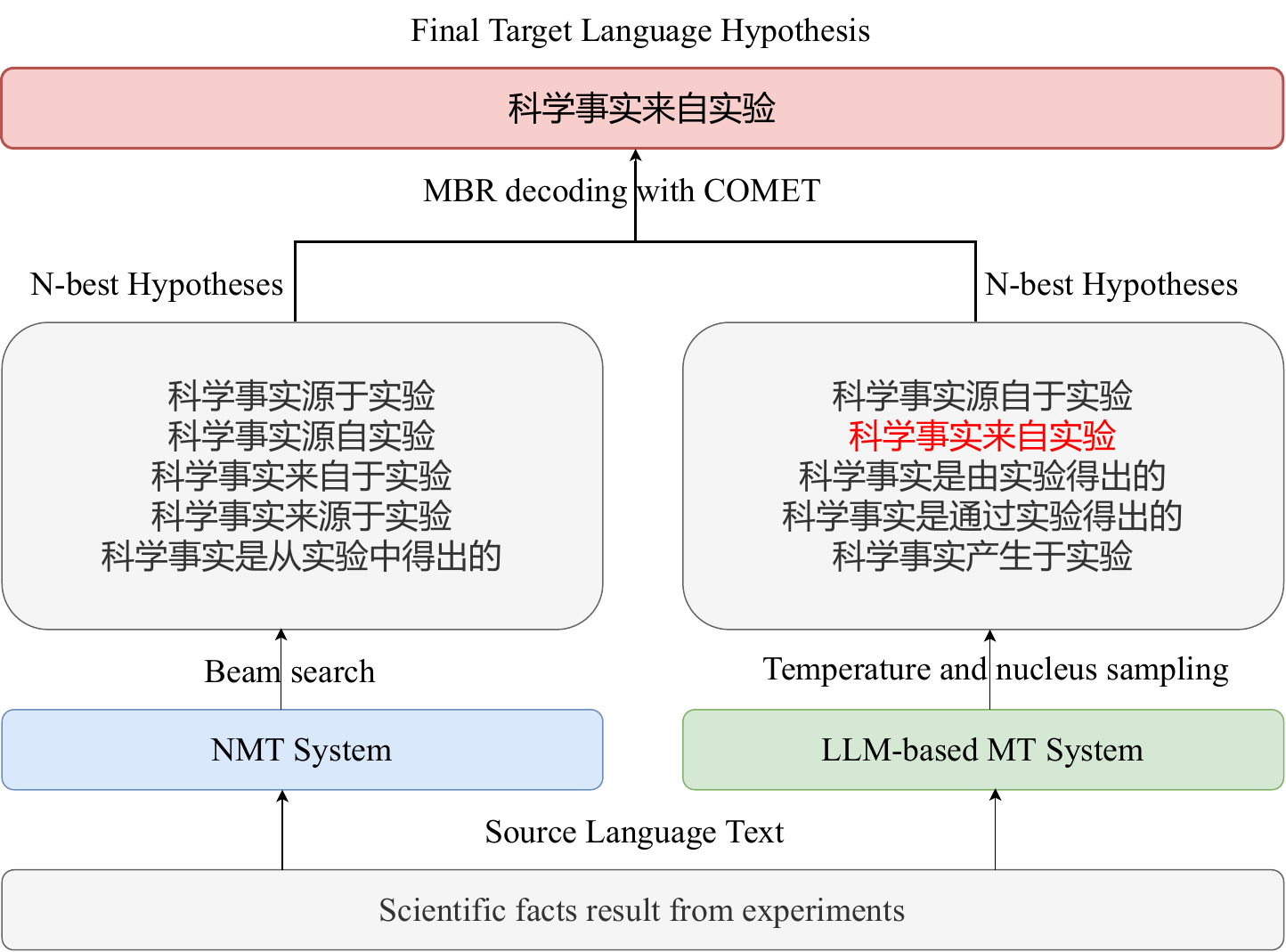}
\caption{\centering Choose the Final Translation from NMT and LLM hypotheses Using MBR Decoding.}
\label{MBR}
\end{figure*}

\subsection{Supervised Fine-tuning}


LLMs have shown remarkable performance on a wide range of NLP tasks by leveraging in-context learning \cite{brown2020language}. However, this approach exhibits several drawbacks: performance is highly dependent on the quality of examples \cite{vilar2023prompting}, outputs are plagued by overgeneration \cite{bawden2023investigating}, and inference cost are greatly increased by processing all input pairs. When parallel data is available, LLMs can perform supervised fine-tuning (SFT) on translation instructions \cite{li2024eliciting}. Drawing inspiration from the recognized significance of data quality in other applications \cite{zhou2024lima},we use the cometkiwi model \cite{rei2022cometkiwi} to filter out large amounts of high-quality parallel data. Here, we use efficient lightweight low-rank adaptation (LoRA) fine-tuning, where we apply LoRA to all modules of feed-forward network.

\subsection{Contrastive Preference Optimization}

Contrastive Preference Optimization (CPO) \cite{xu2024contrastive} aims to mitigate two fundamental shortcomings of SFT. First, SFT’s methodology of minimizing the discrepancy between predicted outputs and gold-standard references inherently caps model performance at the quality level of the training data. This limitation is significant, as even human-written data, traditionally considered high-quality, is not immune to quality issues. Secondly, SFT lacks a mechanism to prevent the model from rejecting mistakes in translations. While strong translation models can produce high-quality translations, they occasionally exhibit minor errors, such as omitting parts of the translation. Preventing the production of these near-perfect but ultimately flawed translation is essential. To overcome these issues, we introduce CPO to train the LLM-based MT model using specially curated triplet preference data. Here, we construct a high-quality preference data for the WMT24 general MT task, and like the SFT stage, only update the weights of the added LoRA parameters.

\subsection{Minimum Bayes Risk Decoding}

Minimum Bayesian Risk (MBR) \cite{kumar2004minimum,eikema2020map} decoding aims to find the output that maximizes the expected utility function, which measures the similarity between the hypothesis and the reference. For MT, this could be an automated evaluation metric such as COMET \cite{rei2020comet}. Garcia et al. \cite{garcia2023unreasonable} train their own language models, sample multiple hypotheses and choose a final translation using MBR decoding, which has been shown to improve the translation capabilities of task-specific models \cite{fernandes2022quality}. Subsequently, Farinhas et al. \cite{farinhas2023empirical} find that MBR is also suitable for LLM-based MT. They provide a comprehensive study on ensembling translation hypotheses, proving that MBR decoding is a very effective method and can improve translation quality using a small number of samples. As shown in Figure \ref{MBR}, we simultaneously collect the N-best translations generated by the NMT system based on beam search and the N-best translations generated by the LLM-based MT system based on temperature and nucleus sampling (with t=0.8 and p=0.95), and then use MBR Decoding selects the final translation.

\section{Experiment}

\subsection{Setup}
We use the open-source fairseq \cite{ott2019fairseq} to train NMT models, and then use SacreBLEU \citep{post-2018-call}\footnote{\url{https://github.com/mjpost/sacrebleu}} and wmt20-comet-da model \cite{rei2020comet} to measure system performance. The main parameters are as follows: each model is trained using 8 GPUs, batch size is 6144, parameter update frequency is 2, and learning rate is 5e-4. The number of warmup steps is 4000, and model is saved every 1000 steps. The architecture we used is described in section 3.1. We adopt dropout, and the rate varies across different training phases. R-Drop is used in model training, and we set $\lambda$ to 5.


We use Llama2-13B as the backbone model of our LLM-based MT system. In our three-stage training process, the first stage uses full fine-tuning, and the last two stages use LoRA fine-tuning. If LoRA is used, lora\_rank is 32, lora\_alpha is 64, lora\_dropout is 0.05, and lora\_modules are "q\_proj", "v\_proj", "k\_proj", "o\_proj", "gate\_proj", "down\_proj", "up\_proj". Furthermore, in the first and third stages, we use open-source ALMA \footnote{\url{https://github.com/fe1ixxu/ALMA}} for training, while in the second stage, we use open-source llama-recipes \footnote{\url{https://github.com/meta-llama/llama-recipes}} for training. The parameters during training are the default configurations of the corresponding codes.

\subsection{Results}

Tables \ref{result} shows the evaluation results of en$\rightarrow$zh NMT systems and LLM-based MT systems on WMT23 general test sets. On NMT systems, we use BiT and R-Drop to build a strong baseline, then use DD, FT and ST for data enhancement, and use AT and CL for more efficient training, and finally use TEL to ensemble multiple models ability. On LLM-based MT systems, we use CPT and SFT to build a strong baseline, and use CPO for further optimization. To ensemble two different types of translation systems, we use MBR decoding to select the final translation, which has been shown to be better than MBR decoding of a single translation system in terms of COMET scores.

\begin{table}[ht]
\large
\begin{center}
\begin{adjustbox}{width=\columnwidth,center}
\begin{tabular}{@{}lcc@{}}
\hline
WMT23 general test set & BLEU & COMET \\
\hline
NMT baseline (BiT \& R-Drop) & 54.24 & 0.6289 \\
+ DD, FT \& ST  & 56.33 & 0.6580 \\
+ AT &  57.03 & 0.6648 \\
+ CL & 58.58 &  0.6830 \\
+ TEL  & \textbf{59.34} & 0.6928 \\
+ NMT MBR & 58.88  &  0.7178 \\
\hline
LLM-based MT baseline (CPT \& SFT) & 52.18 & 0.6553 \\
+ CPO  &  53.09 & 0.6907 \\
+ LLM-based MT MBR & 52.16  &  0.7102 \\
\hline
+ NMT \& LLM-based MT MBR & 56.41  &  \textbf{0.7234} \\
\hline
\end{tabular}
\end{adjustbox}
\caption{BLEU and COMET scores of en$\rightarrow$zh NMT systems and LLM-based MT systems.}\label{result}%
\end{center}
\end{table}

\subsection{Pre-processing and Post-processing}

On the WMT24 general test set, we observe that there are some emoticons and URLs in the source text. To prevent the model from translating them incorrectly, we replace the emoticons and URLs with ”Do Not Translate“ (DNT) labels in pre-processing, and then restore the DNT labels back in post-processing. By doing so, we can reduce some translation errors for emoticons and URLs.

\section{Conclusion}

This paper presents the submission of HW-TSC to the WMT24 general MT Task. On the one hand, we use training strategies such as R-Drop, BiT, DD, FT, BT, AT, CL, and TEL to train the NMT system based on the deep Transformer-big architecture. On the other hand, we use CPT, SFT, and CPO to train the LLM-based MT system. Finally, we use MBR decoding to select the final translation result from the hypotheses generated by these two systems. By using these enhancement strategies, our submission achieved a competitive result in the final evaluation. Relevant experimental results also demonstrate the effectiveness of our strategies. 

\bibliography{emnlp2023}
\bibliographystyle{acl_natbib}

\end{document}